\documentclass{article}

     \PassOptionsToPackage{numbers, compress}{natbib}

\usepackage{hyperref}       
\usepackage{booktabs}       

\usepackage{algorithm,algorithmic}
\usepackage{makecell}
\usepackage{multirow}
\usepackage{amssymb, multirow, paralist, color,amsmath,amsthm}
\newtheorem{thm}{Theorem}

\newtheorem{lemma}{Lemma}

\newtheorem{assumption}{Assumption}
\usepackage{enumitem}
\usepackage{textcase}
\usepackage[T1]{fontenc}
\usepackage{mathtools}
\usepackage{cleveref}
  
\usepackage{empheq}
\usepackage{color,tikz}
\definecolor{darkgreen}{rgb}{0.00,0.5,0.00}
\usepackage{fancybox}

\newenvironment{mybox}[1]{%
    \begin{tcolorbox}[title={~~#1}]%
    }{
    \end{tcolorbox}
}

\definecolor{myteal}{RGB}{27,158,119}
\definecolor{myorange}{RGB}{217,95,2}
\definecolor{myred}{RGB}{231,41,138}
\definecolor{mypurple}{RGB}{152,78,163}
\definecolor{myblue}{rgb}{.9, .9, 1}
\definecolor{mygreen}{RGB}{0,100,0}
\definecolor{mycyan}{rgb}{0.88,1,1}
\definecolor{mydarkred}{RGB}{192,47,25}

\def \S {\mathbf{S}}
\def \A {\mathcal{A}}

\def \R {\mathbb{R}}

\def \w {\mathbf{w}}
\def \v {\mathbf{v}}
\def \t {\mathbf{t}}
\def \x {\mathbf{x}}

\def \E {\mathbb{E}}

\def \x {\mathbf{x}}
\def \p {\mathbf{p}}

\def \1 {\mathbf{1}}

\usepackage[skins]{tcolorbox}
\def \z {\mathbf{z}}
\def \s {\mathbf{s}}

\def \u {\mathbf{u}}

\def \I {\mathbb{I}}
\def \P {\mathcal{P}}
\def \Q {\mathcal{Q}}

\usepackage{cancel}
\usepackage{colortbl}
\usepackage[colorinlistoftodos,bordercolor=orange,backgroundcolor=orange!20,linecolor=orange,textsize=normalsize]{todonotes}

\def \B {\mathcalB}

\def \O {\mathcal O}

\usepackage{booktabs}
\usepackage{tablefootnote}

\newlength\mytemplen
\newsavebox\mytempbox

\makeatletter
\newcommand\mybluebox{%
	\@ifnextchar[
	{\@mybluebox}%
	{\@mybluebox[0pt]}}

\def\@mybluebox[#1]{%
	\@ifnextchar[
	{\@@mybluebox[#1]}%
	{\@@mybluebox[#1][0pt]}}

\def\@@mybluebox[#1][#2]#3{
	\sbox\mytempbox{#3}%
	\mytemplen\ht\mytempbox
	\advance\mytemplen #1\relax
	\ht\mytempbox\mytemplen
	\mytemplen\dp\mytempbox
	\advance\mytemplen #2\relax
	\dp\mytempbox\mytemplen
	\colorbox{myblue}{\hspace{1em}\usebox{\mytempbox}\hspace{1em}}}

\def \E {\mathbb{E}}

\def \x {\mathbf{x}}

\def \D {\mathcal{D}}
\def \z {\mathbf{z}}
\def \u {\mathbf{u}}

\def \w {\mathbf{w}}

\def \R {\mathbb{R}}
\def \S {\mathcal{S}}

\def \Q {\mathcal{Q}}

\def \A {\mathcal{A}}
\def \q {\mathbf{q}}
\def \v {\mathbf{v}}

\def \p {\mathbf{p}}
\def \q {\mathbf{q}}

\usepackage{amssymb}
\usepackage{pifont}

\def \B {\mathcal{B}}

\def \s {\mathbf{s}}

\def \t {\mathbf{t}}

\def \P {\mathcal{P}}

     \usepackage[final]{neurips_2021}

\title{Algorithmic Foundations of Empirical X-risk Minimization}

\author{%
  Tianbao Yang \\
Department of Computer Science and Engineering\\
  Texas A\&M University, College Station, TX 77843 \\
  \texttt{tianbao-yang@tamu.edu}
}

\begin{document}

\maketitle

\begin{abstract}
This manuscript introduces a new optimization framework for machine learning and AI, named {\bf empirical X-risk minimization (EXM)}. X-risk is a term introduced to represent a family of compositional measures or objectives, in which each data point is compared with a large number of items explicitly or implicitly for defining a risk function. It includes surrogate objectives of many widely used measures and non-decomposable losses, e.g., AUROC, AUPRC, partial AUROC, NDCG, MAP,  precision/recall at top $K$ positions, precision at a certain recall level, listwise losses, p-norm push, top push, global contrastive losses, etc. While these non-decomposable objectives and their optimization algorithms have been studied in the literature of machine learning, computer vision, information retrieval, and etc,  optimizing these objectives  has encountered some unique challenges for deep learning. In this paper, we present recent rigorous efforts for EXM with a focus on its algorithmic foundations and its applications. We introduce a class of algorithmic techniques for solving EXM with smooth non-convex objectives.  We formulate EXM into three special families of non-convex optimization problems belonging to  non-convex compositional optimization, non-convex min-max optimization and non-convex bilevel optimization, respectively. For each family of problems, we present some strong baseline algorithms and their complexities, which will motivate further research for improving the existing results. Discussions about the presented results and future studies are given at the end.  Efficient algorithms for optimizing a variety of X-risks are implemented in the LibAUC library at \url{www.libauc.org}. 
\end{abstract}
	
\section{Introduction}
A widely studied fundamental problem in machine learning (ML) is the so-called empirical risk minimization (ERM)~\cite{books/daglib/0033642},  i.e., 
\begin{align}\label{eqn:erm}
	\min_{\w\in\R^d}F(\w):=\frac{1}{n}\sum_{i=1}^n\ell(\w; \x_i),
\end{align}
where $\w$ denotes the model parameter,  $\x_i$ denotes an individual data point, and $\ell(\w; \x)$ denotes a loss of the model on the data point $\x$. Tremendous studies have been devoted to this problem.  Based on the assumption that the functional value and gradient for each loss $\ell(\w, \x_i)$ can be easily computed, sovling the ERM problem enjoys a nice property, i.e.,  an \underline{U}nbiased \underline{S}tochastic \underline{G}radient can be easily calculated based on one random data point or a mini-batch of randomly sampled data points, which is referred to as the USG property in this paper.

However, ERM cannot cover many interesting measures/objectives where the individual loss and its gradient for each data cannot be easily computed, rendering the USG property fail with a mini-batch of sampled data points. Examples  include not only many widely used traditional measures such as average precision (AP) and normalized discounted cumulative gain (NDCG), but also include emerging objectives for AI, e.g., the contrastive loss for {\it self-supervised learning}~\cite{sogclr}. Nevertheless, the large-scale optimization of these measures/objectives for deep learning is still an emerging field. 

The major goal of this paper is to revive the large-scale optimization for a family of compositional measures/objectives called X-risks by summarizing our recent efforts. We will discuss definitions, objectives and optimization problems for a variety of X-risks, and present strong baselines for solving the corresponding optimization problems. A formal definition of X-risk is given below. 
    \begin{mybox}{Definition}
       {\large\bf X-risk} refers to a family of compositional measures in which the loss function of each data point is defined in a way that contrasts the data point with a large number of items.
    \end{mybox}
Mathematically, empirical X-risk minimization (EXM) can be cast into the following abstract optimization problem: 
\begin{align}\label{eqn:xrisk} 
	\min_{\w\in\R^d}F(\w) = \frac{1}{m}\sum_{i=1}^mf_i(g(\w; \x_i, \S_i)),
\end{align}
where $g:\mathcal W\mapsto\mathcal{R}$ is a mapping, $f_i:\mathcal{R} \mapsto \R$, $\S=\{\x_i, \ldots, \x_m\}$ denotes a target set of data points, and $\S_i$ denotes a reference set of items pertinent to $\x_i$. A common form  of the inner function $g(\w; \x_i, \S_i)$ is given below:
\begin{align*}
     g(\w; \x_i, \S_i) = \frac{1}{|\S_i|}\sum_{\z_j\in\S_i}\ell(\w; \x_i, \z_j),
\end{align*}
where we abuse the notation $\ell(; \x_i, \z_j)$ to denote the pair-level loss function.  We refer to the objective function in~(\ref{eqn:xrisk}) as X-risk, which is  to emphasize that the individual term $f_i(g(\w; \x_i, \S_i))$ usually involves the interaction (e.g., comparison) between each individual data $\x_i$ and each item in $\S_i$.  The letter X in ``X-risk" means the cross-coupling between two data in contrast to the parallelism between any two data points in the traditional empirical risk~(\ref{eqn:erm}). We will also generalize the averaged form of $g(\w; \x, \S)$ into an optimization form such that is covers other interesting objectives. 


{\bf Why EXM matters?} There are several reasons to study EXM and revive efficient stochastic algorithms for optimizing various X-risks. First, X-risks cover a broad family of widely used measures/objectives, which include but are not limited to four interconnected categories, i.e., areas under the curves, ranking measures/objectives, performance at the top, and contrastive objectives. Optimizing these measures/objectives have been studied in the literature~\cite{arxiv.2203.15046,DBLP:journals/corr/abs-2006-06889,DBLP:journals/corr/abs-2012-03173, JMLR:v10:rudin09b,AdaRank,NIPS2009_b3967a0e,SVM-NDCG,SoftRank,ApproxNDCG,PiRank,NeuralNDCG,ListNet,ListMLE,Mohapatra2018EfficientOF,brown2020smooth,Cakir_2019_CVPR,Rolinek_2020_CVPR,DBLP:journals/corr/abs-2108-11179,Li2014TopRO,JMLR:v10:rudin09b,DBLP:conf/nips/BoydCMR12,Joachims,pmlr-v119-hiranandani20a,Agarwal2011TheIP,DBLP:journals/corr/abs-2006-12293,goldberger2004neighbourhood,simclrv1,clip}. These objectives have wide applications in modeling a wide range of problems or tasks, e.g., imbalanced data, ranking problems, self-supervised learning. Hence, it is important to develop efficient back-propagation based stochastic optimization algorithms to optimize X-risks for deep learning. Second, most previous works for deep learning~\cite{DBLP:journals/corr/abs-2006-12293,brown2020smooth,Cakir_2019_CVPR,Rolinek_2020_CVPR,DBLP:journals/corr/abs-2108-11179,simclrv1,clip}  simply apply SGD/Adam to the surrogate losses computed based on a mini-batch of samples and need to employ a {\bf large batch size} to achieve a satisfactory performance due to the challenges discussed below.

{\bf What are the Challenges?} The outer function $f_i$ is usually non-linear. In addition, the inner function $g(\w; \x_i, \S_i)$ could be defined explicitly with a finite-sum structure or defined implicitly through a lower-level optimization problem.  Together, it makes the USG property fail even with a mini-batch of samples from $\S$ and $\S_i$. Hence, existing algorithms (e.g., SGD, Adam) and their analysis that rely on the USG property of the full gradient will fail to converge. 

\begin{figure}[t]
\begin{center}
\centerline{\includegraphics[width=\columnwidth]{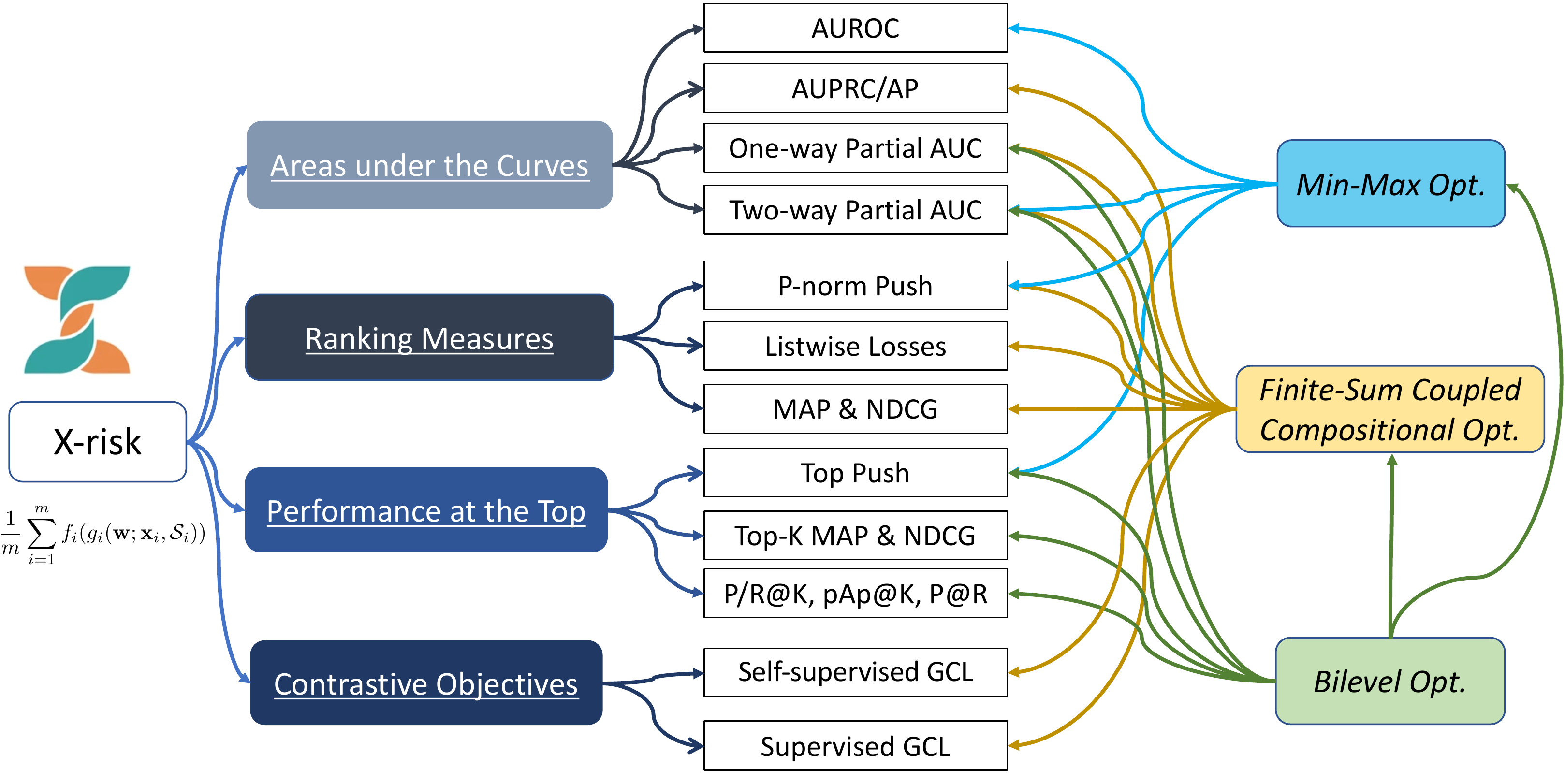}}
\vspace{-0.1in}
\caption{Mappings of four different families of X-risks into different optimization problems. Definitions of each type of X-risk is given in Section~\ref{sec:def}. The relationships between these different X-risks are illustrated in Figure~\ref{fig:graph}. Arrows mean special cases.}
\label{fig:map}
\end{center}
\end{figure}


{\bf What makes it possible and what are the benefits of our algorithms?} Depending on convexity of the outer function $f$ and the structure of the inner function $g$, we  map EXM  into three special families of non-convex optimization problems, namely,   non-convex compositional optimization, non-convex min-max optimization and non-convex bilevel optimization. The recent advances of large-scale stochastic optimization~\cite{rafique2018non,yan2020sharp,DBLP:journals/corr/abs-2006-06889,lin2019gradient,2020arXiv200713605I,DBLP:journals/corr/abs-2008-08170,yang2020global,DBLP:journals/corr/abs-2112-05604,wang2017stochastic,wang2016accelerating,Ghadimi2020AST,Yuan2019EfficientSN,Zhang2019ASC,qi2021online,chen2021solving,Zhang2021MultiLevelCS,balasubramanian2020stochastic,jiang2022optimal,Zhang2020OptimalAF,DBLP:journals/corr/abs-2010-07962,DBLP:journals/corr/abs-2102-04671,99401,DBLP:journals/corr/abs-2007-05170,DBLP:journals/corr/abs-2112-04660,khanduri2021a,chen2021closing,DBLP:journals/corr/abs-2110-07004,arxiv.2104.14840,guo2021randomized} (an extended discussion of these works is given in next section),  including our own works for these non-convex problems~\cite{rafique2018non,yan2020sharp,liu2019stochastic,zhubenchmark,qi2021online,qi2021stochastic,otpaUC,paUCyao,sogclr,qiu2022largescale,arxiv.2104.14840,guo2021randomized},  enable us to optimize a variety of X-risks for deep learning efficiently. The benefits of our algorithms for solving EXM include: (i) they have {\bf provable convergence} with state-of-the-art complexities; (ii) they are {efficient and practical {\bf without requiring a large-batch size}; (iii) they are easy to implement {\bf with minimal mosifications} on existing codes.


\section{Recent Related Works}
Below, we first review our recent works on optimizing the four categories of X-risks and then review some studies (including ours) for solving the three families of non-convex optimization problems. When we consider ``strong" baseline algorithms in this paper, we keep two criteria in mind, namely practicability and theoretical guarantee. A strong baseline algorithm has to be practical (i.e., simple enough such that it can be easily implemented and exhibits a strong practical performance as well), and has strong theoretical guarantee (i.e., convergence rate or sample complexity.)

\paragraph{Optimization of Areas under the Curves.}
While AUROC maximization has been studied in many previous works (please refer to the survey and references therein~\cite{arxiv.2203.15046}), Liu et al.~\cite{liu2019stochastic} is the first to study the large-scale deep AUROC maximization algorithms. They formulate the problem into a non-convex min-max optimization problem and present two stochastic algorithms following the proximal-point framework proposed in~\cite{liu2019stochastic}. The algorithm is refined in~\cite{DBLP:journals/corr/abs-2006-06889} and employed in~\cite{DBLP:journals/corr/abs-2012-03173} for deep AUROC maximization on large-scale medical image datasets. Comparison between different objectives for deep AUROC maximization is studied in~\cite{zhubenchmark}. Large-scale deep AUPRC maximization by stochastic algorithms with {\it convergence guarantee} is first studied in~\cite{qi2021stochastic}. The authors formulate the problem into a novel finite-sum coupled compositional optimization problem, and propose the first algorithm for AP maximization with a convergence guarantee without using a large batch size. The algorithm and theory is later refined and improved in~\cite{DBLP:journals/corr/abs-2107-01173,wangsox}. 

Recently, provable deep partial AUROC (pAUC) maximization is studied in~\cite{otpaUC} and \cite{paUCyao}. The former work focuses on optimizing one-way pAUC with false positive rate less than some value and two-way pAUC. They have formulated one-way partial AUC maximization  as weakly convex optimization and finite-sum coupled compositional optimization, and formulated two-way partial AUC maximization as weakly convex min-max optimization and three-level finite-sum coupled compositional optimization.  The work~\cite{paUCyao} focuses on optimizing one-way partial AUC with false positive rate in a range $(\alpha, \beta)$ and formulates the problem into non-smooth difference-of-convex programming problems. In this paper, we also present a {\it unified bilevel optimization} framework for both one-way pAUC and two-way pAUC regardless the range of false positive rate. 

\paragraph{Optimization of Ranking Measures.}
While optimization of different ranking measures has been considered in many previous works~\cite{JMLR:v10:rudin09b,AdaRank,NIPS2009_b3967a0e,SVM-NDCG,SoftRank,ApproxNDCG,PiRank,NeuralNDCG,ListNet,ListMLE,Mohapatra2018EfficientOF}, most of them are not scalable to big data or applicable to deep learning with a large number of items to be ranked. Some recent works have considered some heuristic tricks (e.g. using a large batch size) for deep learning~\cite{brown2020smooth,Cakir_2019_CVPR,Rolinek_2020_CVPR,DBLP:journals/corr/abs-2108-11179}, which are not practical.   Recently, Qiu et al.~\cite{qiu2022largescale} consider optimization of ranking measures for deep learning, including MAP/NDCG, top-$K$ MAP/NDCG, and listwise losses (e.g., ListNet). The optimization of MAP/NDCG and ListNet is formulated as finite-sum coupled compositional optimization due to their similarities to AP maximization.  For top-$K$ MAP/NDCG maximization, the authors formulate the problem into a novel multi-block bilevel optimization, and have presented convergence analysis for the algorithm to converge to a stationary point.

\paragraph{Optimization of Performance at the Top.}
Performance at the top refers to a set of performance measures that are computed based on examples which are ranked at the top. From this perspective, top-$K$ MAP/NDCG and pAUC are also a kind of performance at the top. Other measures of similar kind include accuracy at the top such as precision at top-$K$ positions (P@$K$), recall at top-$K$ positions (R@$K$), and precision at a certain recall level (P@R). The top-push objective~\cite{Agarwal2011TheIP,Li2014TopRO} is another example of performance at the top, which is an extreme case of p-norm push~\cite{JMLR:v10:rudin09b} and a special case of one-way pAUC.  The pAp@$K$ metric proposed in~\cite{Budhiraja2020RichItemRF} is also an example of performance at the top, which measures the
probability of correctly ranking a top-ranked positive instance over top-ranked negative instances. Optimization of these performance at the top measures for learning a linear/kernel model has been considered in the literature~\cite{DBLP:conf/nips/BoydCMR12,Joachims,pmlr-v119-hiranandani20a,Agarwal2011TheIP,Li2014TopRO}, which are not adept at learning deep neural networks. 
Recently, Adam et al.~\cite{DBLP:journals/corr/abs-2006-12293} have considered optimizing various accuracy at the top measures for deep learning. However, the do not provide any algorithms with convergence guarantee. This is addressed in this work by formulating these problems as bilevel optimization problems due to their similarities to top-$K$ MAP/NDCG maximization~\cite{qiu2022largescale}.  

\paragraph{Optimization of Contrastive Objectives.}
Contrastive objectives have been optimized by using the gradient descent method~\cite{goldberger2004neighbourhood}, which is not suited for deep learning with big data. A straightforward stochastic method is to update the model parameter based on the gradient of a mini-batch contrastive objective that is defined by a mini-batch of samples~\cite{simclrv1,clip}. However, this naive approach does not guarantee convergence and usually requires a large batch size for finding an accurate solution~\cite{simclrv1}. Recently, Yuan et al.~\cite{sogclr} develop a solution for addressing this issue for self-supervised contrastive learning of visual representations. Instead of working with mini-batch contrastive loss, they directly optimize a global contrastive loss by formulating it as finite-sum coupled compositional optimization. Then they develop a stochastic algorithm based on the finite-sum coupled compositional optimization theory. 

\paragraph{Non-Convex Concave Min-Max Optimization.} Non-convex concave min-max optimization is pioneered by~\cite{rafique2018non}. Following this seminal work,  a bulk of studies have proposed different stochastic algorithms for non-convex min-max optimization with convergence guarantee. These algorithms can be categorized into two classes, namely proximal-point based methods~\cite{rafique2018non,yan2020sharp,liu2019stochastic,DBLP:journals/corr/abs-2006-06889}, single-loop methods~\cite{lin2019gradient,guo2021stochastic,2020arXiv200713605I,DBLP:journals/corr/abs-2008-08170,yang2020global,DBLP:journals/corr/abs-2112-05604}. A strong baseline of proximal-point based methods was developed in~\cite{rafique2018non} for non-convex concave problems with a complexity of $O(1/\epsilon^6)$ for finding an $\epsilon$-level nearly stationary point for weakly-convex concave min-max problems. A strong baseline of proximal-point methods for weakly-convex strongly concave min-max problem is proposed in~\cite{yan2020sharp} with a complexity of $O(1/\epsilon^4)$ and  is then refined and improved in~\cite{DBLP:journals/corr/abs-2006-06889} with better complexities under strong conditions.  A strong baseline of single-loop stochastic methods is proposed in~\cite{guo2021stochastic} with a complexity of $O(1/\epsilon^4)$ for solving smooth non-convex strongly concave min-max problems without using a large batch size. Improved complexities have been established for more  complex algorithms~\cite{yang2020global,DBLP:journals/corr/abs-2112-05604,DBLP:journals/corr/abs-2008-08170}. 

\paragraph{Non-Convex Compositional Optimization.}
Stochastic compositional optimization (SCO) refers to a family of optimization problems, in which the objective function is a two-level or multi-level nested and possibly expected-value functions~\cite{wang2017stochastic,Yang2019MultilevelSG}. In the literature, various algorithms have been proposed for two-level SCO~\cite{wang2017stochastic, wang2016accelerating,Ghadimi2020AST,Yuan2019EfficientSN,Zhang2019ASC,qi2021online,chen2021solving,Zhang2020OptimalAF} and multi-level SCO~\cite{Zhang2021MultiLevelCS,balasubramanian2020stochastic,chen2021solving,jiang2022optimal,Zhang2020OptimalAF}. However, these algorithms are not suitable for solving the finite-sum couple compositional optimization problems. 

Recently, a novel class of SCO was studied, which is referred to as finite-sum coupled compositional optimization (FCCO)~\cite{wangsox}.  Its objective is of the form $\frac{1}{n}\sum_{i=1}^n f_{\xi_i}(\mathbb E_{\omega|\xi_i}[g_\omega(\mathbf w; \xi_i)])$. The support for outer level random variable $\xi$ is finite.  The finite-sum structure makes it possible to develop more practical algorithms without relying on huge batch size per-iteration as in previous work~\cite{hu2020biased}. It was first studied in~\cite{qi2021stochastic} for maximizing the point-estimator of area under precision-recall curve known as average precision (AP). Recently, it was comprehensively studied in~\cite{wangsox} and more applications of FCCO have been demonstrated in ML/AI.

\paragraph{Non-Convex Bilevel Optimization.} Although bilevel optimization has a long history in the literature~\citep{Colson07anoverview,doi:10.1080/10556780802102586,Kunisch13abilevel,pmlr-v119-liu20l,doi:10.1137/16M105592X,Shaban2019TruncatedBF}, non-asymptotic convergence was not established until recently~\cite{99401,DBLP:journals/corr/abs-2010-07962,DBLP:journals/corr/abs-2007-05170,DBLP:journals/corr/abs-2102-04671,DBLP:journals/corr/abs-2112-04660,khanduri2021a,chen2021closing,DBLP:journals/corr/abs-2110-07004}, which propose different algorithms for solving stochastic bilevel optimization problems. 
However, these  studies do not explicitly consider the challenge for dealing with multi-block bilevel optimization problems. Guo et al.~\cite{guo2021randomized} is the first that studies the multi-block bilevel optimization and propose a variance-reduced algorithm. Nevertheless, the algorithm proposed in~\cite{guo2021randomized} aims for the improved sample complexity and is less practical. 

Another challenge for tackling the bilevel optimization problems arising in EXM is that the lower level problem is non-smooth and non-strongly convex. Although the issue of non-strong convexity of the lower level problem has been tackled in some prior works~\cite{pmlr-v119-liu20l,arxiv.2106.07991,arxiv.2203.01123}, their algorithms are not practical for solving EXM. In this work, we present a strong baseline algorithm developed by Qiu et al.~\cite{qiu2022largescale}. For the special lower level problem arising in maximizing top-K NDCG, they smooth the lower-level objective and also add a small quadratic term to make the objective strongly convex such that the optimal solution does not change too much. For handling multiple blocks, their algorithm only needs to update the variables for the sampled blocks.  

\paragraph{Other Related Works.} There are other efforts trying to unify the optimization of different metrics/measures.  Adam et al.~\cite{DBLP:journals/corr/abs-2006-12293} consider a unified objective function, which is a linear combination of false positive rate and false negative rate for a specific threshold that is a function of a group of prediction scores. The function for defining the threshold is typically the score of one example  in the group satisfying a certain criterion. In particular, they consider the function to be the k-th largest scores in a group of scores, which covers several metrics, e.g., Precision at a certain Recall level, Recall at top $K$ positions, etc.  However, as mentioned before they do not provide any convergence guarantee for the stochastic algorithms that they developed. In contrast, our algorithms are directly applicable to all measures they have considered and come with convergence guarantee.  Narasimhan et al.~\cite{10.5555/3454287.3455251} have considered a unified objective that is a function of accuracy of different groups of data, which covers G-mean, H-mean, Q-mean, F-measure, AUPRC, etc. They transform the problem into constrained optimization problems. However, the theoretical convergence guarantee is usually developed for convex models (e.g., linear model), and is not applicable to deep learning.  

\paragraph{Other X-risks.} Finally, we note that the techniques presented in this paper are applicable to many other X-risks  beyond those mentioned above and discussed later. In~\cite{wangsox} the authors have discussed many other interesting applications of FCCO in ML/AI, e.g., deep survival analysis, deep latent variable models, softmax losses, and model agnostic meta-learning. In addition, some other problems can be formulated as X-risk optimization as well, including some distributionally robust optimization problems~\cite{qi2021online,DBLP:journals/corr/abs-2012-06951}, graph neural network optimization~\cite{DBLP:conf/icml/GraphFM}, multi-class cross-entropy loss with many classes.

\section{Three Classes of Optimization Problems}\label{sec:def}
In this section, we introduce three classes of non-convex optimization problems, namely  finite-sum coupled compositional optimization, multi-block min-max optimization and multi-block bi-level optimization, which are useful for formulating EXM.

\paragraph{\bf Finite-sum Coupled Compositional Optimization (FCCO)} 
Many X-risks will be formulated as a special family of compositional optimization problems named FCCO~\cite{wangsox}:
\begin{align}\label{eqn:fcco}
    \min_{\w}\frac{1}{m}\sum_{i=1}^mf_i\left(\E_{\z|\x_i}[g(\w; \x_i, \z)]\right), 
\end{align}
where  $f_i$ is a simple deterministic function that is not necessarily convex. 
It is clear that~(\ref{eqn:xrisk}) reduces to FCCO when  $g(\w; \x_i, \S_i)=\frac{1}{|\S_i|}\sum_{\z\in\S_i}\ell(\w; \x_i, \z)$ and $f_i$ is a simple deterministic function. 

\paragraph{\bf Multi-block Min-Max  Optimization (MBMMO)}
 The min-max objective is usually a reformulation of the X-risk in~(\ref{eqn:xrisk}). When $f_i$ is convex and $g(\w; \x_i, \S_i)$ has a finite-sum structure. Let $f_i^*$ be the convex conjugate  of $f_i$ and $g(\w; \z_i, \S_i)=\frac{1}{|\S_i|}\sum_{\z'\in\S_i}\ell(\w; \z_i, \z')$. Then, we can formulate~(\ref{eqn:xrisk}) into the following: 
\begin{align}
    \min_{\w}\max_{\s\in\Omega^m} \frac{1}{m}\sum_{i=1}^m s_i^{\top}\frac{1}{|\S_i|}\sum_{\z\in\S_i}\ell(\w; \x_i, \z) - f_i^*(s_i). 
\end{align} 
This min-max structure may enable one to develop faster algorithms given the vast literature on min-max optimization. More generally, we consider the following multi-block min-max optimization:  
\begin{align}\label{eqn:minmaxdec}
    \min_{\w}\max_{\s\in\Omega^m} \frac{1}{m}\sum_{i=1}^m F_i(\w, \s_i),
\end{align}
where $\s=(\s_1, \ldots, \s_m)$. 
It is notable that this is slightly different from the standard finite-sum structure considered in the literature for the minimax objective~\cite{rafique2018non,yang2020global},  which refers to that $F(\w,\s)=\frac{1}{m}\sum_{i=1}^mF_i(\w, \s)$.

\paragraph{\bf Multi-block Bilevel Optimization (MBBO).}
As a generalization of both FCCO and MBMMO, we will consider multi-block bilevel optimization (MBBO)  given by 
\begin{equation}\label{eqn:bim}
\begin{aligned}
    \min_{\w}&\quad \sum_{i=1}^mF_i(\w, \s_i(\w)),\quad \text{where}\quad\s_i(\w)= \arg\min_{\s\in\Omega} L_i(\w, \s), i=1, \ldots, m.
\end{aligned}
\end{equation}
EXM (\ref{eqn:xrisk}) becomes MBBO when  $g(\w; \z_i, \S_i)$ is formulated in a way that depends on the optimal solution of a lower-level problem.

Before ending this section, we would like to point out that both 
 the FCCO problem~(\ref{eqn:fcco}) and the MBMMO problem~(\ref{eqn:minmaxdec}) can be formulated as special cases of MBBO~(\ref{eqn:bim}).  Finally, we present a mapping of various X-risks into  different families of non-convex optimization problems in Figure~\ref{fig:map}. It is notable that the mapping for some X-risks might not be unique indicating that there are multiple ways to optimize them. Details will be provided in next section.

\section{Definitions, Formulations and Mappings of X-risks}
In this section, we review the definitions of a variety of X-risks and their formulations, and present their mappings into different optimization problems. 

\subsection{Areas under the Curves}\label{sec:auc}
In this subsection, we consider binary classification or bipartite ranking problems.  Let $\z = (\x, y)$ denote an input-output pair, where $\x\in\mathcal X $ denotes the input data and $y\in\{1,-1\}$ denotes its class label. Let $\P_+$ denote the distribution of positive examples and $\P_-$ denote the distribution of negative examples.  Let $h_\w(\x):\mathcal X\rightarrow \R$ denote a predictive function parameterized by a vector $\w\in\R^d$. Let $\ell(\w; \x, \x') = \ell(h_\w(\x') - h_\w(\x))$ denote a pairwise surrogate loss for a positive-negative pair $(\x, \x')$. Let $\mathbb I(\cdot)$ be an indicator function of a predicate, and $[n]=\{1,\ldots, n\}$. For a given threshold $t$,  the true positive rate (TPR) can be written as $\text{TPR}(t) = \Pr(h(\x)>t|y = 1) = \E_{\x\sim \P_+}[\I(h(\x)>t)]$, and the false positive rate (FPR) can be written as $\text{FPR}(t) = \Pr(h(\x)>t | y=-1) = \E_{\x\sim\P_-}[\I(h(\x)>t)]$

For a set of training examples $\S=\{(\x_i, y_i), i\in[n]\}$, let $\S_+$ and $\S_-$ be the subsets of $\S$ with only positive  and negative examples, respectively, and let $n_+=|\S_+|$ and $n_-=|\S_-|$ be the number of positive and negative examples, respectively. Denote by  $\S^{\downarrow}[k_1,k_2]\subseteq\S$ the subset of examples whose rank in terms of their prediction scores in the descending order are in the range of $[k_1, k_2]$, where $k_1\leq k_2$. Similarly, let $\S^{\uparrow}[k_1,k_2]\subseteq\S$ denote the subset of examples whose rank in terms of their prediction scores in the ascending order are in the range of $[k_1, k_2]$, where $k_1\leq k_2$. 
We denote by $\E_{\x\sim\S}$  the average over $\x\in\S$. Let $D(\p, \mathbf q)$ denote the KL divergence between two probability vectors. Let $\Delta$ denote a simplex of a proper dimension, and $\Pi_{\Omega}[\cdot]$ denote the standard Euclidean projection onto a set $\Omega$. Below, we let $\x\sim\P$ and $\x\sim\S$ denote a random sample from a distribution and a finite set, respectively.

\subsubsection{AUROC} The definition of AUROC (area under ROC curve, a.k.a AUC) gives the following formula for AUC of a predictive function $h_\w$~\cite{Hanley1982}:
\begin{align}\label{eqn:aucd2}
\text{AUC}(h_\w) = \Pr(h_\w(\x_+)> h_\w(\x_-)) = \E_{\x_+\sim\P_+, \x_-\sim\P_-}[\mathbb I(h_\w(\x_+)>h_\w(\x_-))]. 
\end{align}
Using a pairwise surrogate loss, we have the following standard objective for AUROC optimization: 
\begin{align}\label{eqn:eaucd}
\min_{\w\in\R^d} \frac{1}{n_+}\frac{1}{n_-}\sum_{\x_i\in\S_+}\sum_{\x_j\in\S_-}\ell(h_\w(\x_j) - h_\w(\x_i)). 
\end{align}
This can be regarded as a special case of~(\ref{eqn:xrisk}) by setting $g(\w; \x_i, \S_-) = \frac{1}{n_-}\sum_{\x_j\in\S_-}\ell(h_\w(\x_j) - h_\w(\x_i))$ and $f_i(\cdot)=\cdot$. Indeed, this is the simplest form of X-risk as an unbiased stochastic gradient can be easily computed based on a pair of data points consisting of a random positive and a random negative data point.

In the literature, min-max objectives have been proposed for AUROC maximization~\cite{ying2016stochastic,DBLP:journals/corr/abs-2012-03173}, which can be written as 
\begin{equation}\label{eqn:aucminmax}
\begin{aligned}
\min_{\w\in\R^d, (a,b)\in\R^2}\quad\max_{s\in\R}\quad &\E_{\x_i\sim\S_+}[(h_\w(\x_i)-a)^2] + \E_{\x_j\sim\S_-}[(h_\w(\x_j)-b)^2] \\
 & + s(\E_{\x_j\sim\S_-}[h_\w(\x_j)] -\E_{\x_i\sim\S_+}[h_\w(\x_i)]+c)  - \ell^*(s),
\end{aligned}
\end{equation}
where $\ell^*$ is the convex conjugate of a surrogate loss $\ell(\cdot)$ and $c>0$ is a margin parameter. This is a special case of non-convex MMO. 

When we consider multi-task or multi-label AUROC maximization, we can formulate the problem as 
\begin{equation}\label{eqn:aucmbmmo}
\begin{aligned}
\min_{\w\in\R^d, \mathbf a\in\R^K, \mathbf b\in\R^{K}}\quad\max_{\s\in\R^K}\quad &\sum_{k=1}^K\E_{\x_i\sim\S^k_+}[(h_\w(\x_i;k)-a_k)^2] + \E_{\x_j\sim\S^k_-}[(h_\w(\x_j;k)-b_k)^2] \\
 & + s_k(\E_{\x_j\sim\S_-^k}[h_\w(\x_j;k)] -\E_{\x_i\sim\S_+^k}[h_\w(\x_i;k)]+c)  - \ell^*(s_k),
\end{aligned}
\end{equation}
where $k$ denotes the index of tasks or labels. This is a special case of MBMMO.

\subsubsection{AUPRC}  Area under Precision-Recall curve (AUPRC) is an average of the precision weighted by the probability of a given threshold, which can be expressed as
\begin{align*}
\text{AUPRC} = \int_{-\infty}^{\infty}\Pr(y=1|h(\x)\geq c) d\Pr(h(\x)\leq c|y=1),
\end{align*}
where $\Pr(y=1|h(\x)\geq c)$ is the precision at the threshold value of $c$.  For a set of training examples $\S=\S_+\cup\S_-$, a non-parametric estimator of AUPRC is average precision (AP): 
\begin{align}
\label{eqn:OnlAvgPre-p}
    \text{AP}&=\frac{1}{n_{+}} \sum\limits_{\x_i\in\S_+} \frac{\sum\limits_{\x_j\in\S_+}\mathbb I(h(\x_j) \geq h(\x_i))}{\sum\limits_{\x_j\in\S} \mathbb I(h(\x_j)\geq h(\x_i))}.
\end{align}
It can be shown that AP is an unbiased estimator of AUPRC in the limit $n\rightarrow \infty$~\cite{10.1007/978-3-642-40994-3_29}.

By using a differentiable surrogate loss $\ell(h(\x_j) - h(\x_i))$ in place of $\mathbb I(h(\x_j)\geq h(\x_i))$, we have the following optimization problem:
\begin{align}
    \min_{\w} \frac{1}{n_{+}} \sum\limits_{\x_i\in\S_+} f(\mathbf g(\w; \x_i, \S_+, \S)),
\end{align}
where we define $g_1(\w; \x_i, \S_+)=\sum\limits_{\x_j\in\S_+}\ell(h_\w(\x_j) - h_\w(\x_i))$, $g_2(\w; \x_i, \S)=\sum\limits_{\x_j\in\S} \ell(h_\w(\x_j)- h_\w(\x_i))$, $\mathbf g(\w; \x_i, \S_+, \S)=[g_1(\w; \x_i, \S_+), g_2(\w; \x_i, \S)]$ and $f(\mathbf g)=-\frac{[\mathbf g]_1}{[\mathbf g]_2}$. This is a special case of FCCO. 

\subsubsection{Partial AUROC (pAUC)} 
There are two variants of pAUC, namely one-way pAUC and two-way pAUC. We first give the mathematical definitions and then present objectives for their optimization. 

One-way pAUC is defined as the area under the ROC curve with FPR restricted in the range $[\alpha, \beta]$, which can be expressed by 
\begin{align}\label{eqn:paucd1}
\text{pAUC}(h, \alpha, \beta) =\Pr(h(\x_+)> h(\x_-), h(\x_-)\in[\text{FPR}^{-1}(\beta), \text{FPR}^{-1}(\alpha)]), 
\end{align}
where $\text{FPR}^{-1}(u) = \inf\{t\in\R: \text{FPR}(t)\leq u\}$ denotes the $u$ quantile of $f(\x_-), \x_-\sim \P_-$. Hence, one-way pAUC maximization can be formulated as
\begin{align}\label{eqn:epaucd2}
\min_{\w} \frac{1}{n_+}\frac{1}{n_-}\sum_{\x_i\in\S_+}\sum_{\x_j\in\S^\downarrow_-[k_1+1, k_2]}\ell(h_\w(\x_j)-h_\w(\x_i)). 
\end{align}
where $k_1=\lceil n_-\alpha\rceil, k_2 = \lfloor n_-\beta \rfloor$. A challenge for solving the above problem is that  the selection of top ranked negative examples from $\S_-$, i.e., $\S_-^\downarrow[k_1+1, k_2]$ for some fixed $k_1, k_2$ is non-differentiable. To tackle this challenge, we present three ways to transform the above problem into a form that facilitates the design of stochastic algorithms. 

Two-way pAUC is defined as the area under the ROC curve with FPR restricted in a range of $[0, \beta]$ and TPR restricted in a range of $[\alpha, 1]$, which can be expressed as 
\begin{align}\label{eqn:tpaucd}
\text{TPAUC}(h, \alpha, \beta) = \Pr(h(\x_+)> h(\x_-), h(\x_-)\geq  \text{FPR}^{-1}(\beta), h(\x_+)\leq \text{TPR}^{-1}(\alpha)\}). 
\end{align}
It inspires the following optimization problem with a surrogate loss $\ell(\cdot)$:
\begin{align}\label{eqn:etpaucd2}
\min_{\w}\frac{1}{n_+}\frac{1}{n_-}\sum_{\x_i\in\S_+^{\uparrow}[1, k_1]}\sum_{\x_j\in\S^\downarrow_-[1, k_2]}\ell(h_\w(\x_j) - h_\w(\x_i))
\end{align}
where  $k_1=\lfloor n_+\alpha\rfloor, k_2 = \lfloor n_-\beta \rfloor$. 

{\bf Weakly Convex Optimization.} The first transform is given by~\cite{otpaUC} which relies on an assumption that $\ell(\cdot)$ is non-decreasing (a common assumption)  and $\alpha=0$. 
When $\ell(\cdot)$ is non-decreasing and $\alpha=0$, then the problem~(\ref{eqn:epaucd2}) is equivalent to 
\begin{align}\label{eqn:opauccvar} 
\hspace*{-0.1in}\min_{\w}\min_{\s\in\R^{n_+}} F(\w, \s)= \frac{1}{n_+}\sum_{\x_i\in\S_+} \left(s_i  +  \frac{1}{\beta} \psi_i(\w, s_i)\right),
\end{align}
where $\psi_i(\w, s_i) = \frac{1}{n_-} \sum_{\x_j\in \S_-}(\ell(\w; \x_i, \x_j) - s_i)_+$. The new objective function $F(\w, \s)$ is $\rho$-weakly convex in terms of $(\w, \s)$ for some $\rho>0$ if $\ell(\w; \x_i, \x_j)$ is a smooth function of $\w$. 
Similarly for two-way pAUC maximization, 
\cite{otpaUC} shows that the problem~(\ref{eqn:etpaucd2}) is equivalent to 
\begin{align}\label{eqn:tpAUC}
    \min_{\w, q'\in\R, \q\in\R^{n_+}}\max_{\s\in[0,1]^{n_+}}q' +\frac{1}{n_+\alpha}\sum_{\x_i\in\S_+}s_i(q_i+\frac{1}{\beta}\psi_i(\w; q_i) - q'), 
\end{align}
where $\psi_i(\w, q_i) = \frac{1}{n_-} \sum_{\x_j\in \S_-}(\ell(\w; \x_i, \x_j) - q_i)_+$. The new objective function is weakly convex in terms of $(\w, q', \q)$ and is concave in terms $\s$ if $\ell(\w; \x_i, \x_j)$ is a smooth function of $\w$.

{\bf Compositional Optimization.} The second way is to define a soft estimator of pAUC, which is formulated as a compositional optimiztion problem. We consider the DRO estimators of pAUC under the condition that $\ell(\cdot)$ is non-decreasing, which are proposed in~\cite{otpaUC}. In particular, for one-way pAUC maximization one can consider the following problem: 
\begin{align}\label{eqn:pauceskl}
\min_{\w}\frac{1}{n_+}\sum_{\x_i\in\S_+}\lambda\log \E_{\x_j\in\S_-}\exp(\frac{\ell(\w; \x_i, \x_j)}{\lambda}).
\end{align}
For two-way pAUC maximization we consider the following problem: 
\begin{align}\label{eqn:tpauceskl}
\min_{\w} f_1 (\frac{1}{n_+}\sum_{\x_i\in\S_+}f_2(g_i(\w))),
\end{align}
where $f_1(s) = \lambda'\log (s)$, $f_2(g) = g^{\lambda/\lambda'}$ and $g_i(\w) = \E_{\x_j\sim\S_-}\exp(\ell(\w; \x_i, \x_j)/\lambda) $. It is notable that~(\ref{eqn:pauceskl}) is a special case of FCCO~(\ref{eqn:fcco}), while~(\ref{eqn:tpauceskl}) is slightly more complicated with a three-level composition. 

{\bf Bilevel Optimization.} The above transformation relies on the condition that $\ell(\cdot)$ is non-decreasing, which does not cover all surrogate losses (e.g., square loss, barrier hinge loss, etc). To eschew this issue, we propose a bilevel optimization formulation for~(\ref{eqn:epaucd2}) and~(\ref{eqn:etpaucd2}) inspired by~\cite{qiu2022largescale}. The key is the following lemma. 
\begin{lemma}\label{lemma:1}\cite{qiu2022largescale}
Let $\s\in\R^n$, $\epsilon\in(0,1)$ and $\lambda_*= \arg\min_{\lambda}(k+\epsilon)\lambda +\sum_{i=1}^n(s_i -\lambda)_+$, then $\lambda_*=s_{[k+1]}$, where $s_{[k]}$ denotes the $k$-th largest element in $\s$.
\end{lemma}
Base on the above problem, we can transform a top-$K$ selection operator for selecting a data point whose score is in the top-$K$ positions among all scores $h_\w(\x), \forall\x\in\S$, i.e., $\I(\x\in\S_{[1,K]})$, into $\I(h_\w(\x)> \lambda(\w))$, where 
\begin{align}
    \lambda(\w)= \arg\min_{\lambda\in\R}\frac{(K+\epsilon)}{|\S|}\lambda +\frac{1}{|\S|}\sum_{\x\in\S}(h_\w(\x) -\lambda)_+.
\end{align}
However, one challenge for dealing with this optimization problem in a bilevel optimization framework is that $\lambda(\w)$ is non-smooth and non-Lipschitz continuous function of $\w$. To address this issue, we can smooth the objective function and add a small quadratic term to make it strongly convex, i.e., considering
\begin{align}\label{eqn:topKs}
    \hat\lambda(\w)= \arg\min_{\lambda\in\R}L(\lambda, \w; K, \S):=\frac{K+\epsilon}{|\S|}\lambda+ \frac{\tau_2}{2}\lambda^2+\frac{1}{|\S|}\sum_{\x\in\S}\tau_1\ln(1+\exp((h_\w(\x)-\lambda)/\tau_1)),
\end{align}
where $\tau_1,\tau_2$ are small constants. 
The following lemma justifies the above smoothing.
\vspace{-0mm}
\begin{lemma}\label{lemma:4}\cite{qiu2022largescale}
Assuming $h_\w(\x)$ is smooth and bounded, if $\tau_1=\tau_2=\varepsilon$ for some $\varepsilon\ll 1$ , then we have $|\hat\lambda(\w) - \lambda(\w)|\leq O(\varepsilon)$ for any $\w$. In addition,  $L(\lambda, \w; K, \S)$ is twice-differentiable and  smooth and strongly convex in terms of $\lambda$ for any $\w$. 
\vspace{-0mm}
\end{lemma}
Hence, below we will use the optimization problem~(\ref{eqn:topKs}) for constructing a bilevel optimization problem. 

In particular, we transform (\ref{eqn:epaucd2}) into  
\begin{align*}
\min_{\w} &  \frac{1}{n_+}\frac{1}{n_-}\sum_{\x_i\in\S_+}\sum_{\x_j\in\S_-}\I(h_\w(\x_j)>\lambda_2(\w))\I(h_\w(\x_j)<\lambda_1(\w))\ell(h_\w(\x_j)-h_\w(\x_i))\\
\lambda_1(\w)&=\arg\min_{\lambda\in\R} L(\lambda, \w; k_1-1, \S_-), \quad 
\lambda_2(\w)=\arg\min_{\lambda\in\R} L(\lambda, \w; k_2, \S_-). 
\end{align*}
By replacing the indicator function $\I(s>0)$ with a non-decreasing differentiable surrogate function $\phi(s)$, we have the following optimization problem for one-way pAUC maximization: 
\begin{equation}\label{eqn:opauc-bi}
\begin{aligned}
\min_{\w} \quad&  \frac{1}{n_+}\frac{1}{n_-}\sum_{\x_i\in\S_+}\sum_{\x_j\in\S_-}\phi(h_\w(\x_j) -\lambda_2(\w))\phi(\lambda_1(\w)-h_\w(\x_j))\ell(h_\w(\x_j)-h_\w(\x_i))\\
\lambda_1(\w)&=\arg\min L(\lambda, \w; k_1-1, \S_-), \quad 
\lambda_2(\w)=\arg\min L(\lambda, \w; k_2, \S_-). 
\end{aligned}
\end{equation}
Similarly, for two-way pAUC maximization~(\ref{eqn:etpaucd2}) we can formulate the following bilevel optimization:
\begin{equation}
\begin{aligned}
\min_{\w} \quad&  \frac{1}{n_+}\frac{1}{n_-}\sum_{\x_i\in\S_+}\sum_{\x_j\in\S_-}\phi(h_\w(\x_j) -\lambda_2(\w))\phi(\lambda_1(\w)-h_\w(\x_i))\ell(h_\w(\x_j)-h_\w(\x_i))\\
\lambda_1(\w)&=\arg\min_{\lambda\in\R}  L(\lambda, \w; n_+-k_1, \S_+), \quad 
\lambda_2(\w)=\arg\min_{\lambda\in\R}  L(\lambda, \w; k_2, \S_-). 
\end{aligned}
\end{equation}

\subsection{Ranking Measures/Losses}
We first introduce some notations. Let $\Q$ denote the query set of size $N$, and $q\in\Q$ denote a query. $\S_q$ denotes a set of $N_q$ items (e.g., documents, movies) to be ranked for $q$. For each $\x^q_i\in\S_q$, let $y^q_i\in\R^+$ denote its relevance score, which measures the relevance between query $q$ and item $\x^q_i$. Let $\S^+_q\subseteq\S_q$ denote a set of $N^+_q$ items \emph{relevant} to $q$, whose relevance scores are \emph{non-zero}. Denoted by $\S=\{(q, \x^q_i),q\in\Q, \x^q_i\in\S^+_q\}$ all relevant query-item (Q-I) pairs. 
Let $h_\w(\x, q)$ denote the predictive function for $\x$ with respect to the query $q$, whose parameters are denoted by $\w\in\R^d$ (e.g., a deep neural network). Let
\begin{align*}
r(\w; \x, \S_q) = \sum_{\x'\in\S_q}\I(h_\w(\x', q) - h_\w(\x, q)\geq 0)
\end{align*}
denote the rank of $\x$ with respect to the set $\S_q$, where we simply ignore the tie.

\paragraph{MAP and NDCG.}
According to the definition of NDCG, the averaged NDCG over all queries can be expressed by  
\begin{align*} 
\text{NDCG:}\quad \frac{1}{N}\sum_{q=1}^N\frac{1}{Z_q}\sum_{\x_i^q\in \S^+_q} \frac{2^{y^q_i}-1}{\log_2(r(\w; \x^q_i, \S_q)+1) },
\end{align*}
where $Z_q$ is the maximum DCG of a perfect ranking of items in $\S_q$, which can be pre-computed. 
Note that $\x^q_i$ are summed over $\S_q^+$ instead of $\S_q$, because only relevant items have non-zero relevance scores and contribute to NDCG.

MAP can be similarly computed as
 \begin{align*} 
\text{MAP:}\quad\frac{1}{N}\sum_{q=1}^N\sum_{\x_i^q\in \S^+_q} \frac{r(\w; \x^q_i, \S^+_q)}{r(\w; \x^q_i, \S_q) }.
\end{align*} 

By replacing $r(\w; \x, \S_q)$ with a surrogate function $g(\w; \x, \S_q)=\sum_{\x'\in\S_q}\ell(h_\w(\x', q) - h_\w(\x, q))$, we can formulate both NDCG and MAP maximization as FCCO, i.e., 
 \begin{align*} 
\min_{\w}\frac{1}{|\S|}\sum_{(q, \x_i^q)\in\S}f_{q,i}(g(\w; \x_i^q, S_q)).
\end{align*}

\paragraph{Top-$K$ MAP and Top-$K$ NDCG.}
The top-$K$ variant of NDCG or MAP only computes the corresponding score for those that are ranked in the top-$K$ positions. For example, the top-$K$ NDCG is defined as:
\begin{align*} 
\frac{1}{N}\sum_{q=1}^N\frac{1}{Z^K_q}\sum_{\x_i^q\in\S^+_q}\I(\x_i^q\in\S_q[K]) \frac{2^{y^q_i}-1}{\log_2(r(\w; \x^q_i, \S_q)+1) },
\end{align*} 
where $\S_q[K]$ denotes the top-$K$ items whose prediction scores are in the top-$K$ positions among all items in $\S_q$, and $Z^K_q$ denotes the top-$K$ DCG score of the perfect ranking. 

To handle the top-$K$ selection, we can formulate it as a bilevel optimization problem siminlar as before, i.e., 
\begin{align*} 
\min_{\w}\quad & - \frac{1}{N}\sum_{q=1}^N\frac{1}{Z^K_q}\sum_{\x_i^q\in\S^+_q}\I(h_q(\x_i^q; \w) - \lambda_q(\w)> 0) \frac{2^{y^q_i}-1}{\log_2(g(\w; \x^q_i, \S_q)+1) },\\
\lambda_q(\w) & = \arg\min_{\lambda} L(\lambda, \w; K, \S_q), \forall q\in\Q.
\end{align*} 
By replacing the indicator function with a differentiable surrogate function $\phi(\cdot)$, the above problem reduces to the following form
\begin{align*} 
\min_{\w}\quad &  \frac{1}{|\S|}\sum_{(q,\x_i^q)\in\S}\phi(h_q(\x_i^q; \w) - \lambda_q(\w)) f_{q,i}(g(\w; \x_i^q, \S_q)),\\
\lambda_q(\w) & = \arg\min_{\lambda} L(\lambda, \w; K, \S_q), \forall q\in\Q.
\end{align*} 
The above prolem is a special case of MBBO. 

\paragraph{Listwise Ranking Losses.} One can also optimize other listwise losses, e.g., ListNet~\cite{ListNet}, ListMLE~\cite{ListMLE}. 
For example,  the objective function of ListNet can be  defined by a cross-entropy loss between two probabilities of list of scores:
\begin{align*}
	&\min_{\w} -\sum_{q}\sum_{\x^q_i\in\S_q}P(y^q_i)\log \frac{\exp(h_\w(\x^q_i; q)}{\sum_{\x\in\S_q}\exp(h_\w(\x^q_i; q))},
	\end{align*}
where $P(y^q_i)$ denotes a probability for a relevance score $y^q_i$ (e.g., $P(y^q_i)\propto y^q_i$). We can map the above problem into FCCO, where $g(\w; \x^q_i, \S_q) = \frac{1}{|\S_q|}\sum_{\x\in\S_q}\exp(h_\w(\x; q)-h_\w(\x^q_i; q))$ and $f_{q, i}(g) = P(y^q_i)\log (g)$.

\paragraph{p-norm Push (Pn-Push).}
For a bipartite ranking or binary classification problem,  a $p$-norm push objective can be defined as~\cite{JMLR:v10:rudin09b}: 
\begin{align*}
	\min_{\w} \frac{1}{|\S_+|}\sum_{\z_i\in\S_+}\left(\frac{1}{|\S_-|}\sum_{\z_j\in\S_-}\ell(h_\w(\z_j) - h_\w(\z_i))\right)^p,
\end{align*}
where $p>1$ and $\ell(\cdot)$ is similar as in the AUC loss.  We can cast this problem into FCCO by defining the inner function as $g(\w; \z_i, \S_-)= \frac{1}{|\S_-|}\sum_{\z_j\in\S_-}\ell(h_\w(\z_j) - h_\w(\z_i))$ and the outer function $f(g)= g^p$. One can also switch the summation over $\S_+$ and the summation over $\S_-$ to define another version of $p$-norm push objective as in the original paper.

\subsection{Performance at the Top}
In this subsection, we consider the binary classification setting as in subsection~\ref{sec:auc}.

\paragraph{Top Push (T-Push).}
The top push objective is defined as 
\begin{align}\label{eqn:tp}
	\min_{\w} \frac{1}{|\S_+|}\sum_{\z_i\in\S_+}\ell(\max_{\z_j\in\S_-}h_\w(\z_j) - h_\w(\z_i)). 
\end{align}
This form is difficult to work with. There are several ways to transform this problem into one of the considered optimization problems. \cite{Li2014TopRO} presents a transformation that formulate the above problem as a min-max problem assuming $\ell$ is convex and non-decreasing. However, it introduces non-decomposable constraints on the dual variable. 

We can use simpler transforms. First, we note that the above objective is equivalent to the surrogate objective of pAUC in~(\ref{eqn:epaucd2}) with $k_1=0$ and $k_2=1$.  Then, we can reduce the problem into weakly convex minimization problem~(\ref{eqn:opauccvar}) when $\ell(\cdot)$ is non-decreasing, or reduce the problem into a bilevel optimization problem in~(\ref{eqn:opauc-bi}). 


\paragraph{Precision/Recall at Top $K$ Positions (P@$K$, R@$K$).}
For binary classification, the recall at top $K$ positions is defined as the recall among those examples whose predictions scores are in the top $K$ positions.  Leveraging Lemma~\ref{lemma:1} and Lemma~\ref{lemma:4}, we can easily formulate the problem of maximizing recall at top $K$ positions as a bilevel optimization problem: 
\begin{equation}\label{eqn:recK}
\begin{aligned}
    \max_{\w}\quad&\frac{1}{|\S_+|}\sum_{\z_i\in\S_+}\mathbb I(h_\w(\z_i) > \lambda(\w))\\
    \lambda(\w)&= \arg\min_{\lambda\in\R}L(\lambda, \w; K,\S). 
\end{aligned}
\end{equation}
Prec@K is equivalent to Rec@K up to a constant. 
By replacing the indicator function by a surrogate function, we have the following problem for optimizing Rec@K (Prec@K), 
\begin{equation}\label{eqn:recK}
\begin{aligned}
    \min_{\w}\quad&\frac{1}{|\S_+|}\sum_{\z_i\in\S_+}\ell(\lambda(\w) - h_\w(\z_i))\\
    \lambda(\w)&= \arg\min_{\lambda\in\R}L(\lambda, \w; K,\S), 
\end{aligned}
\end{equation}
where $\ell(\cdot)$ is a differentiable non-decreasing function.  If there are multiple labels or multiple tasks, the problem becomes MMBO. 

Another way is to formulate the Rec@K and Prec@K as a FCCO problem. The idea is to compare the rank of a positive data with $K$, i.e., 
\begin{equation*}
\begin{aligned}
    \max_{\w}\quad&\frac{1}{|\S_+|}\sum_{\z_i\in\S_+}\mathbb I(r(\w; \z_i, S) \leq K)\\
    r(\w; \z_i, \S)&= \sum_{\z_j\in\S}\I(h_\w(\z_j)\geq h_\w(\z_i))). 
\end{aligned}
\end{equation*}
Replacing the indicator function by a surrogate function, we have the following FCCO problem:
\begin{equation}\label{eqn:recK2}
\begin{aligned}
    \min_{\w}\quad&\frac{1}{|\S_+|}\sum_{\z_i\in\S_+}\ell_1\left( \sum_{\z_j\in\S}\ell_2(h_\w(\z_j)- h_\w(\z_i)) - K\right)
    \end{aligned},
\end{equation}
where $\ell_1(\cdot)$ and $\ell_2(\cdot)$ are monotonically non-decreasing  function, e.g., sigmoid function

\paragraph{Precision at a Certain Recall level (P@R).}
The precision at certain recall level $\alpha=K/n_+$ means that if we consider a threshold defined by the $K$-th largest score among all positive examples, what is the proportion of positive examples among those whose scores are higher than the threshold. The precision can be computed as $\frac{K}{K+\text{FP}}$, where $\text{FP}$ denotes the number of false positives whose scores are larger than the $K$-th largest score among all positive examples~\cite{DBLP:journals/corr/abs-2006-12293}. Hence, maximizing precision at the recall level $\alpha$ is equivalent to minimizing the false positive rate at a threshold given by the $K$-th largest score among all positive examples. Thus, we can formulate the problem of maximizing precision at a certain recall level as a bilevel optimization problem:
\begin{equation}\label{eqn:prerec}
\begin{aligned}
    \min_{\w}\quad&\frac{1}{|\S_-|}\sum_{\z_j\in\S_-}\ell( h_\w(\z_j) - \lambda(\w))\\
    \lambda(\w)&= \arg\min_{\lambda\in\R}L(\lambda, \w; K, \S_+), 
\end{aligned}
\end{equation}
where the objective is a surrogate of FPR. If there are multiple labels or multiple tasks, the problem becomes MMBO. 

\paragraph{pAp@$K$.} The pAp@$K$ metric is a measure that combines the
partial-AUC and the precision@$K$ metrics, defined as probability of correctly ranking a top-ranked positive instance over top-ranked negative instances~\cite{Budhiraja2020RichItemRF}. It has been used to evaluate recommendation systems and in real-world deployments. A surrogate loss for pAp@$K$ is given by 
\begin{align}\label{eqn:pAp}
\min_{\w}\frac{1}{k_1K}\sum_{\x_i\in\S_+^{\downarrow}[1, k_1]}\sum_{\x_j\in\S^\downarrow_-[1, K]}\ell(h_\w(\x_j) - h_\w(\x_i))
\end{align}
where  $k_1=\min(K, n_+)$. We can see that the above objective is very similar to that of two-way pAUC~(\ref{eqn:etpaucd2}). Hence, we can similarly formulate the optimization of pAp@$K$ as a bilevel optimization problem. 

Other performance at the top measures can be similarly formulated (please refer to~\cite{DBLP:journals/corr/abs-2006-12293} for definitions).

\subsection{Global Contrastive Losses (GCL)}
The contrastive objectives have long been used for distance metric learning~\cite{goldberger2004neighbourhood} and recently emerge in self-supervised learning methods~\cite{simclrv1,clip}. Below, we discuss one-way self-supervised contrastive loss for vision data, and two-way self-supervised contrastive loss for image-text data, and also present an example of supervised contrastive objectives. We also note that similar contrastive losses have been used for learning with graph data~\cite{yang2021graphformers} and text data~\cite{DBLP:conf/emnlp/KarpukhinOMLWEC20}. 

\paragraph{One-way Self-supervised Contrastive Loss.}  Let $\S=\{\x_1, \ldots, \x_n\}$ denote the set of training images, let $\mathcal P$ denote a set of data augmentation operators that can be applied to each image to generate a copy. Let $\A(\cdot)\in\mathcal P$ denote a random data augmentation operator, and let  $\x\in\S$ denote a random example from $\S$. 
Let $\S_i=\{\A(\x): \forall\A\in\P, \forall\x\in\S\setminus \{\x_i\}\}$ denote all training images including their augmented versions but excluding that of $\x_i$. Let $h_\w(\cdot)$ denote the encoder network parameterized by $\w\in\R^d$ that outputs a normalized feature representation of an input image.  Below, $\A(\x_i)$ and $\A(\x_j)$ denote two independent random data augmentations applied to $\x_i$ and $\x_j$ independently. 

A global contrastive loss is given by~\cite{sogclr}:
\begin{align}\label{eqn:gcl}
   \min_{\w} \E_{\x_i\sim\S, \A, \A'\sim \P} [-\tau \ln\frac{\exp(h_\w(\A(\x_i))^{\top}h_\w(\A'(\x_i))/\tau)}{\varepsilon + g(\w; \x_i, \A, \S_i)}],
\end{align}
where $\varepsilon$ is a small constant and  \begin{align}\label{eqn:g}
  g(\w; \x_i, \A, \S_i) & = \sum_{\z\in\S_i}(\exp(h_\w(\A(\x_i))^{\top}h_\w(\z)/\tau).
\end{align}

\begin{figure}[t]
\begin{center}
\centerline{\includegraphics[width=\columnwidth]{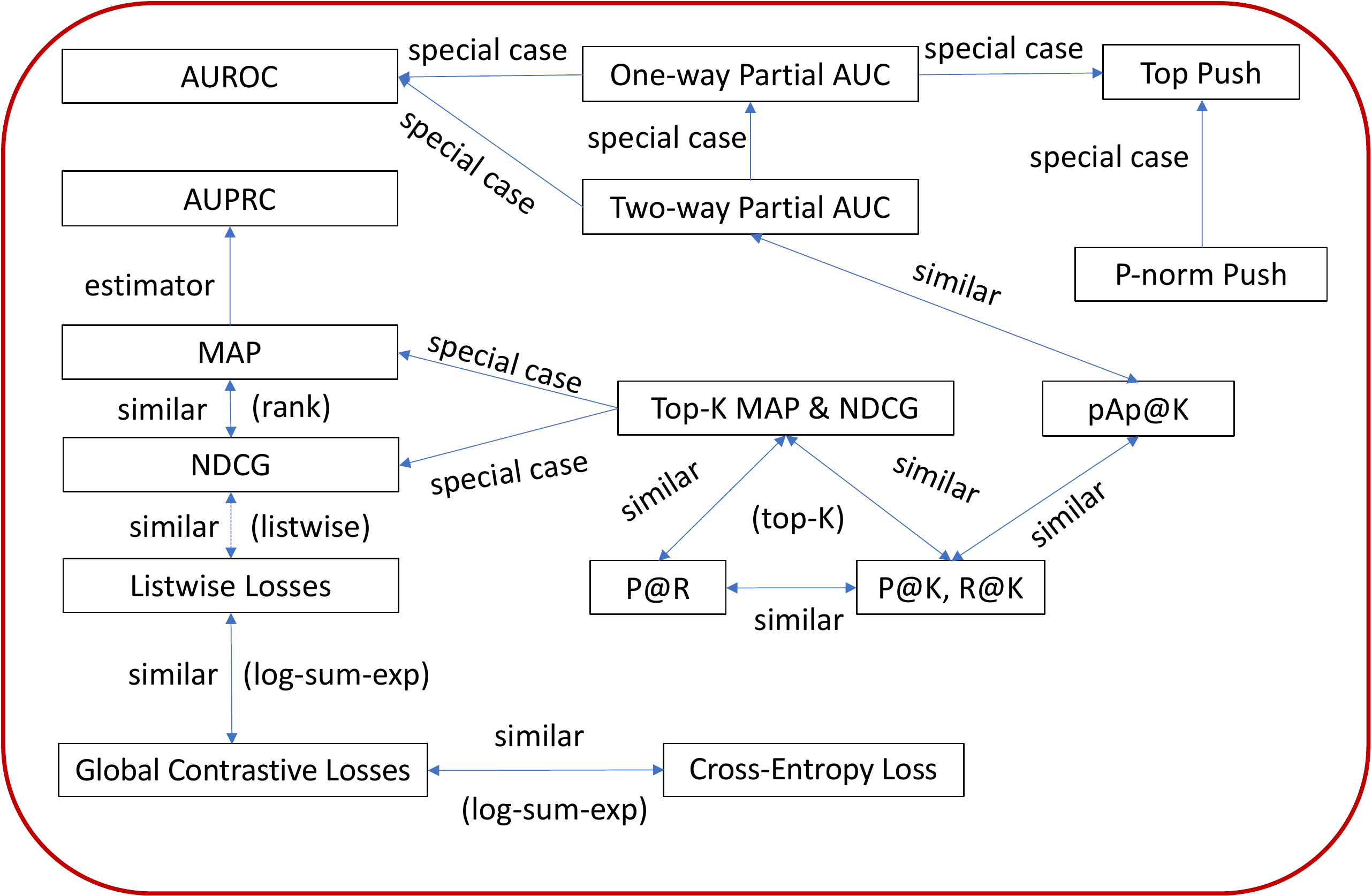}}
\vspace{-0.1in}
\caption{Relationships between different X-risks. AUROC is a special case of one-way pAUC and two-way pAUC. One-way pAUC with FPR in a range $(0, \alpha)$ is a special case of two-way pAUC. Top Push is a special case of one-way pAUC and p-norm push. AP is a non-parametric estimator of AUPRC. MAP and NDCG are similar in the sense that they are functions of ranks. Top-$K$ MAP, top-$K$ NDCG, R@$K$, P@$K$, pAp@$K$, P@R are similar in the sense that they all involve the computation of $K$-th largest scores in a set. Listwise losses, global contrastive losses and cross-entropy losses are similar in the sense that they all involve the sum of log-sum-exp term.}
\label{fig:graph}
\end{center}
\end{figure}
\paragraph{Two-way Self-supervised Contrastive Loss.}  Given a set of image-text pairs $\D=\{(\x_1, \t_1), \ldots, (\x_n, \t_n)\}$. We denote by $h_{\w_1}$ and $h_{\w_2}$ the encoder network for the image data and the text data, respectively.  We can consider optimizing a global two-way contrastive loss: 
\begin{align*}
\min_{\w}  -\frac{\tau}{n}\sum_{i=1}^n\log\frac{\exp(h_{\w_1}(\x_i)^{\top}h_{\w_2}(\t_i)/\tau)}{\sum_{\t\in\D}\exp(h_{\w_1}(\x_i)^{\top}h_{\w_2}(\t)/\tau)}-\frac{\tau}{n}\sum_{i=1}^n\log\frac{\exp(h_{\w_1}(\x_i)^{\top}h_{\w_2}(\t_i)/\tau)}{\sum_{\x\in\D}\exp(h_{\w_1}(\x)^{\top}h_{\w_2}(\t_i)/\tau)}.
\end{align*}
The denominator may exlude the positive data from consideration. It is not difficult to see that both one-way contrastive loss and two-way contrastive loss are special cases of FCCO. 

\paragraph{Supervised Contrastive Learning.}
There are many studies about supervised contrastive learning~\cite{globerson-2006-metric,DBLP:conf/nips/KhoslaTWSTIMLK20}. Given a set of data points $\S=\{(\x_1, y_1), \ldots, (\x_n, y_n)\}$,  let $h_\w(\cdot)$ denote an embedding model to be learned. A supervised contrastive objective  (e.g., NCA) is given by 
\begin{align}
    \min_{\w}-\frac{1}{n}\sum_{\x\in\S}\frac{\sum_{\x'\in\S(\x)}\exp(-\|h_\w(\x) - h_\w(\x')\|^2)}{\sum_{\x'\in\S}\exp(-\|h_\w(\x) - h_\w(\x')\|^2)},
\end{align}
where $\S(\x)$ denotes the set of examples who are in the same class as $\x$. One can also consider other contrastive objectives, e.g., 
\begin{align}
    \min_{\w}-\frac{1}{n}\sum_{\x\in\S}\log \frac{\sum_{\x'\in\S(\x)}\exp(-\|h_\w(\x) - h_\w(\x')\|^2)}{\sum_{\x'\in\S}\exp(-\|h_\w(\x) - h_\w(\x')\|^2)},
\end{align}
and 
\begin{align}
    \min_{\w}-\frac{1}{n}\sum_{\x\in\S}\sum_{\x'\in\S(\x)}\log\frac{\exp(-\|h_\w(\x) - h_\w(\x')\|^2)}{\sum_{\x'\in\S}\exp(-\|h_\w(\x) - h_\w(\x')\|^2)}.
\end{align}
All three objectives are special cases of FCCO. 

Finally, we present a knowledge graph in Figure~\ref{fig:graph} to illustrate the relationships between different measures/objectives discussed above.

\section{Strong Baseline Stochastic Algorithms}
In this section, we present strong baseline stochastic algorithms for solving EXM. For more generality, we consider three families of non-convex optimization, i.e., FCCO, MBMMO and MBBO.

\subsection{Baseline Stochastic Algorithm for FCCO}
The following FCCO has been studied in our prior works~\cite{qi2021stochastic,wangsox}. 
\begin{equation}\label{eqn:FCCO}
\begin{aligned}
    &\min_{\w\in\R^d}\quad F(\w)=\frac{1}{m}\sum_{i\in\S}f_i(g_i(\w; \S_i))\\
    &\text{where } g_i(\w; \S_i) = \frac{1}{|\S_i|}\sum_{\z\in\S_i}\ell(\w; \x_i, \z),
\end{aligned}
\end{equation}
where $m=|\S|$.  Let $g_i(\w; \B_{2,i})= \frac{1}{|\B_{2,i}|}\sum_{\z\in\B_{2,i}}\ell(\w; \x_i, \z)$ denote an unbiased estimator of $g_i(\w; \S_i)$  based on a mini-batch of samples in $\B_{2,i}$. 

{\bf Algorithm and Applicability.} A strong baseline algorithm is presented in Algorithm~\ref{alg:sox}~\cite{wangsox}.  The algorithm is applicable to optimization of AUPRC, AP, p-norm push, the compositional objective for one-way pAUC, NDCG/MAP, ListNet, GCL. 

{\bf Convergence Analysis.} For convergence analysis, we make the following assumptions. 
\begin{assumption}\label{asm:lip}
Regarding the problem~(\ref{eqn:FCCO}), the following conditions hold: 
\begin{itemize}
\item $f_i$ is smooth and Lipschitz continuous; 
\item $g_i(\w; \S_i)$ is differentiable, smooth and Lipschitz continuous; 
\item $\E[\|g_i(\w_t; \B_{2,i}) - g_i(\w_t; \S_i)\|^2]\leq O(\frac{1}{B_2})$ and $\E[\|\nabla g_i(\w_t; \B_{2,i}) - \nabla g_i(\w_t; \S_i)\|^2]\leq O(\frac{1}{B_2})$ for $|\B_{2,i}|=B_2$. 
\end{itemize}
\end{assumption}
\begin{thm}\label{thm:sox}
	Let $|\B^t_1|=B_1, |\B^t_{2,i}|=B_2$. Under Assumptions~\ref{asm:lip}, Algorithm~\ref{alg:sox} (equipped with either momentum and Adam-style update) with  $\gamma_0 = O(B_2\epsilon^2)$, $\gamma_1=O(\min\{B_1,B_2\}\epsilon^2)$, $\beta_1 = 1-\gamma_1$, and $\eta = O\left(\min (\gamma_1, \frac{\gamma_0 B_1}{m})\right)$ can find an $\epsilon$-stationary point in $T=O\left(\max\left\{\frac{m}{B_1B_2\epsilon^4}, \frac{1}{\min\{B_1, B_{2}\}\epsilon^4}\right\}\right)$ iterations. 	
\end{thm}
The proof of the above theorem can be found in~\cite{wangsox}. We can also extend Algorithm~\ref{alg:sox} to solving the three-level finite-sum compositional optimization problem~(\ref{eqn:tpauceskl}) for two-way pAUC maximization with a similar complexity.  We refer the readers to~\cite{otpaUC} for more details.

\begin{algorithm}[t]
	\caption{Stochastic Optimization of X-risk (SOX) for FCCO}
	\begin{algorithmic}[1]
	\STATE Initialize $\w_0$,$\u_0$,$\v_0$,$\eta$, $\beta$, $\gamma$ 
		\FOR{$t=0,\ldots,T$}
		\STATE Draw a batch of  $B_1$ samples $\B^t_1\subset \S$
		\FOR{each $i \in \B^t_1$} 
		\STATE Update the estimator of function value $g_i(\w_{t})$ $$\u^{t+1}_i =  (1-\gamma_0)\u^{t}_i  + \gamma_0 g_i(\w_{t};\B^t_{2,i})\hspace*{0.5in} \diamond\B^t_{2,i}\subset\S_i \text{ is a mini-batch of samples}$$
		\ENDFOR 
		\STATE Update the estimator of gradient $\nabla F(\w_{t})$ $$\v_{t+1} =\beta_1\v_{t} + (1-\beta_1)\frac{1}{B_1}\sum_{i\in\B^t_1}\nabla f_i({{\bf\u}^{t}_i}) \nabla g_i(\w_{t};\B^t_{2,i})$$  
		\STATE Update the model parameter $\w_{t+1} = \w_t - \eta \v_{t+1}$ (or the Adam-style update)
		\ENDFOR
	\end{algorithmic}
	\label{alg:sox}
\end{algorithm}

\subsection{Baseline Stochastic Algorithm for MBMMO}
Recall the following MBMMO:
\begin{align}\label{eq:mbmmo}
    \min_{\w\in\R^d}\max_{\s\in\Omega^m}F(\w, \s)=\frac{1}{m}\sum_{i=1}^mF_i(\w, s_i). 
\end{align}
For simplicity of presentation, we let $\nabla_\w F_i(\w, s ;\B)$ and $\nabla_s F_i(\w, s ;\B)$ denote unbiased stochastic gradients in terms of $\w$ and $s$ based on a mini-batch of samples in $\B$, respectively.

{\bf Algorithm and Applicability.} A strong single-loop baseline algorithm is presented in Algorithm~\ref{alg:pd}. It is an extension of an algorithm in our prior work~\cite{arxiv.2104.14840} and a special case of the multi-block min-max bilevel optimization algorithm presented in~\cite{quanqimbmmo}. The primal variable follows a stochastic momentum update and the dual variable follows a coordinate-wise standard stochastic gradient ascent update. Algorithm 1 is applicable to (multi-label or multi-task) AUROC maximization, p-norm push maximization.

{\bf Convergence Analysis.} For convergence analysis,  we make the following assumption regarding the above problem. 
\begin{assumption}\label{ass:1}Regarding the problem~(\ref{eq:mbmmo}), the following conditions hold: 
\begin{itemize}
\item  $F_i(\w, s_i)$ is smooth in terms of both $\w, s_i$, and is Lipschitz continuous in terms of $\w$.
\item $F_i(\w, \cdot)$ is strongly  concave for any $\w$. 
\item  $\E[\|\nabla_\w F_i(\w, s; \B_{2,i}) - \nabla_\w F_i(\w, s)\|^2]\leq O(\frac{1}{B_2})$ and $\E[\|\nabla_s F_i(\w, s; \B_{2,i}) - \nabla_s F_i(\w, s)\|^2]\leq O(\frac{1}{B_2})$ for $|\B_{2,i}|=B_2$. 
\end{itemize}
\end{assumption}
\begin{thm}
Let $|\B^t_1|=B_1, |\B^t_{2,i}|=B_2$. Under Assumption~\ref{ass:1}, Algorithm~\ref{alg:pd} (equipped with either momentum and Adam-style update) with $\eta_0=\O(B_2\epsilon^2)$, $\gamma_1=\O(\min\{B_1,B_2\}\epsilon^2)$,  $\beta_1 = 1- \gamma_1$,  $\eta=\O(\min\left\{\gamma_1,\frac{B_1\eta_0}{m}\right\})$ can find an $\epsilon$-level solution to the primal objective $\max_{\s\in\Omega^m}F(\w, \s)$ with $T=\O(\max\left\{\frac{1}{\min\{B_1,B_2\}\epsilon^4},\frac{m}{B_1B_2\epsilon^4}\right\})$ iterations. 
\end{thm}
The proof of the above theorem is a special case for the problem considered in~\cite{quanqimbmmo}.

\begin{algorithm}[t]
	\caption{Stochastic Optimization of X-risk (SOX) for MBMMO}\label{alg:pd}
	\begin{algorithmic}[1]
\STATE{Initialize: $\w_0\in \R^d, \s_0 \in\R^{m}, \v_0$}   
\FOR{$t=0, 1, ..., T$}
\STATE Draw a batch of  $B_1$ samples $\B^t_1\subset \S$
\FOR{each $i \in \B^t_1$} 
\STATE {$s_{i,t+1} = \Pi_{\Omega}[s_{i,t} + \eta_0 \nabla_s F_i(\w_t, s_{i,t}; \B^t_{2,i})]$} \hfill $\diamond \B^t_{2,i}$ is a mini-batch of samples
\ENDFOR
\STATE {$\v_{t+1} = \beta_1\v_t  +  (1-\beta_1)\frac{1}{B_1}\sum_{i\in\B_1^t}\nabla_{\w} F_i(\w_{t}, s_{i,t}; \B^t_{2,i}) $} 
\STATE$\w_{t+1} = \w_{t} - \eta\v_{t+1}$ (or use Adam-style update)
\ENDFOR 
\end{algorithmic}
\end{algorithm}

A limitation of the analysis of above algorithm is that it requires the smoothness, which is not applicable to weakly convex concave min-max problems without the smoothness condition of the objective, e.g., the problem~(\ref{eqn:tpAUC}) for two-way pAUC maximization. For solving weakly convex concave min-max problems, the baseline algorithm is following the proximal-point based method proposed in our prior work~\cite{rafique2018non}. However,  to increase the efficiency for two-way pAUC maximization, we need to employ stochastic coordinate updates for both the primal and dual variable. We refer interested readers to~\cite{otpaUC} for the algorithm and analysis.

\subsection{Baseline Stochastic Algorithm for MBBO}
We consider the following MBBO:
\begin{equation}\label{eqn:mbbo}
\begin{aligned}
    \min_{\w\in\R^d}&\quad F(\w)=\frac{1}{m}\sum_{i\in\S}f_i(g_i(\w; \S_i))\phi_i(\w, \lambda_i(\w; \S'_i))\\
   \text{where }& g_i(\w; \S_i) =  \frac{1}{|\S_i|}\sum_{\z\in\S_i}\ell(\w; \z_i, \z)\\
   \text{and }& \lambda_i(\w; \S_i')=\arg\min_{\lambda\in\R}  L_i(\lambda, \w; \S_i')= \frac{1}{|\S_i'|}\sum_{\z\in\S_i'}\psi(\lambda, \w; \z).
\end{aligned}
\end{equation}
Let $g_i(\w; \B_i) = \frac{1}{|\B_i|}\sum_{\z\in\B_i}\ell(\w;\z_i, \z)$ and $L_i(\lambda, \w; \bar\B_i)=\frac{1}{|\bar\B_i|}\sum_{\z\in\bar\B_i}\psi(\lambda, \w; \z)$. 

{\bf Algorithm and Applicability.} An efficient stochastic algorithm is presented in Algorithm~\ref{alg:mbbo}. It was originally proposed in~\cite{qiu2022largescale} for optimizing top-$K$ NDCG. The algorithm reduces to Algorithm~\ref{alg:sox} when $\phi_i(\w, \lambda_i(\w; \S'_i))$ is not present.  When $f_i(g)$ is not present and $L_i(\lambda, \w; \S_i')= - \frac{1}{m}\sum_{i\in\S}\phi_i(\w, \lambda_i)$, the problem reduces to min-max optimization and Algorithm~\ref{alg:mbbo} can be simplified as Algorithm~\ref{alg:pd}.    Algorithm~\ref{alg:mbbo}  is applicable to optimization of top-$K$ NDCG, top-$K$ MAP, pAUC, top-push,  P@$K$, R@$K$, pAp@$K$, and P@R. Except for top-$K$ NDCG and MAP, other measures only have one block $m=1$ unless for multi-task learning.

{\bf Convergence Analysis.} For convergence analysis, we make the following assumptions. 
\begin{assumption}\label{ass:3}
Regarding the problem~\ref{eqn:mbbo}, the following conditions hold:\,
\begin{itemize}
    \item Functions $\phi_i,f_i,g_i,L_i$ are smooth and Lipschitz continuous, respectively for all $i$. 
    \item Function $\psi(\lambda, \w; \z)$ is strongly convex in terms of $\lambda$, and $\nabla^2_{\w\lambda}L_i(\w,\lambda),\nabla^2_{\lambda\lambda}L_i(\w,\lambda)$ are Lipschitz continuous respectively with respect to $(\w,\lambda)$ for all $i$. 
    \item $\phi_i(\cdot)$ and $f_i(\cdot)$ are bounded, $\nabla^2_{\w\lambda}\psi(\lambda,\w; \z)$  and $\nabla^2_{\lambda\lambda}\psi(\lambda, \w;\z)$ are bounded for all $\z$. 
    \item $\E[\|g_i(\w; \B_{2,i}) - g_i(\w; \S_i)\|^2]\leq O(\frac{1}{B_2})$ and $\E[\|\nabla g_i(\w; \B_{2,i}) -\nabla g_i(\w; \S_i)\|^2]\leq O(\frac{1}{B_2})$ for $|\B_{2,i}|=B_2$. 
    \item $\E[\|\nabla_\lambda L_i(\w,\lambda_i;\bar\B_{2,i}) - \nabla_\lambda L_i(\w,\lambda_i)\|^2]\leq O(\frac{1}{B_2})$,  $\E[\|\nabla^2_{\lambda,\lambda} L_i(\w,\lambda_i;\bar\B_{2,i}) - \nabla^2_{\lambda,\lambda} L_i(\w,\lambda_i)\|^2]\leq O(\frac{1}{B_2})$, and $\E[\|\nabla^2_{\w,\lambda} L_i(\w,\lambda_i;\bar\B_{2,i}) - \nabla^2_{\w,\lambda} L_i(\w,\lambda_i)\|^2]\leq O(\frac{1}{B_2})$, for $|\bar\B_{2,i}|=B_2$. 
\end{itemize}
\end{assumption}
\begin{algorithm}[t]
\caption{Stochastic Optimization of X-risk (SOX) for MBBO}\label{alg:mbbo}
\begin{algorithmic}[1]
\STATE Initialize $\w_0,\v_0,\lambda^0, \u^0, s^0,  \beta_1, \gamma_0, \gamma_1, \eta_0,  \eta_1$
\FOR{$t=0,1,\dots,T$}
\STATE Draw a batch of $B_1$ samples $\mathcal B^t_1\subset\S$
\FOR{each $i\in\B^t_1$}
\STATE Update $\u_i^{t+1}=(1-\gamma_0)  \u_{i}^t+ \gamma_0 g_i(\w_t;\B^t_{2,i})$\hfill $\diamond\B^t_{2,i}$ is a mini-batch of samples
\STATE Update $s_i^{t+1}=(1-\gamma_1)s_i^t+\gamma_1\nabla^2_{\lambda\lambda}L_i(\w_t,\lambda_i^t;\bar\B^{t}_{2,i})$\hfill $\diamond\bar\B^t_{2,i}$ is a mini-batch of samples
\STATE Update $\lambda_i^{t+1}=\lambda_{i}^t -  \eta_0\nabla_\lambda L_i(\w_t,\lambda_i^t;\bar\B^{t}_{2,i})$
\ENDFOR
\STATE Compute stochastic estimator $\mathbf m_t$ 
\begin{equation*}
    \begin{aligned}
    \mathbf m_t
    &=\frac{1}{B_1}\sum_{i\in\B^t_1}\bigg\{ \bigg(\nabla_\w\phi_i(\w_t,\lambda_i^t)-\nabla_{\w \lambda}^2 L_i(\w_t,\lambda_i^t;\bar\B_{2,i}) [s_i^{t}]^{-1} \nabla_\lambda \phi_i(\w_t,\lambda_i^t)\bigg)f_i(\u_i^t)\\
    &\quad\quad\quad\quad\quad +\phi_i(\w_t,\lambda_i^t)\nabla g_i(\w_t;\B_{2,i}^t)\nabla f_i(\u_i^t)\bigg\}
    \end{aligned}
\end{equation*}
\STATE $\v_{t+1}=\beta_1\v_t+(1-\beta_1) \mathbf m_t$
\STATE $\w_{t+1}=\w_t-\eta_1 \v_{t+1}$ (or the Adam-style update)
\ENDFOR
\end{algorithmic}
\end{algorithm}

\begin{thm}\label{thm_ana}
Let $|\B^t_1|=B_1, |\B^t_{2,i}|=|\bar\B^t_{2,i}|=B_2$. Under Assumption~\ref{ass:3}, SOX-MBBO (Algorithm~\ref{alg:mbbo} equipped with either momentum or Adam-style update) with  $\gamma_0=\O(B_2\epsilon^2), \gamma_0' = \O(B_2\epsilon^2)$, $\gamma_1 = \O(\min\{B_1,B_2\}\epsilon^2)$, $\beta_1 = 1-\gamma_1$, $\eta_0 = \O\left(\frac{B_1\gamma_0}{m}\right)$, $\eta_1 = \O\left(\min\left \{\frac{B_1 \gamma_0}{m},\gamma_1\right\}\right)$ can find an $\epsilon$-stationary point in $T=\O\left( \max\left\{\frac{m}{B_1B_2\epsilon^4},\frac{1}{\min\{B_1,B_2\}\epsilon^4} \right\}\right)$ iterations. 
\end{thm}
The proof of the above theorem can be found in~\cite{qiu2022largescale}.

\begin{table}[t] 
	\caption{Summary of different algorithms and their applicability to different X-risks.}
	\centering
	\label{tab:1}
	\scalebox{0.9}{\begin{tabular}{lcc}
			\toprule
			Alg.&X-risks&Iteration Complexity\\
			\midrule
			    Algorithm~\ref{alg:sox}&\makecell{AUPRC, Pn-Push, pAUC\\NDCG, MAP, GCL, P@$K$, R@$K$}&$\O\left( \max\left\{\frac{m}{B_1B_2\epsilon^4},\frac{1}{\min\{B_1,B_2\}\epsilon^4} \right\}\right)$\\ 
       				\midrule
           Algorithm~\ref{alg:pd}&AUROC, Pn-Push&$\O\left( \max\left\{\frac{m}{B_1B_2\epsilon^4},\frac{1}{\min\{B_1,B_2\}\epsilon^4} \right\}\right)$\\
			\midrule
			Algorithm~\ref{alg:mbbo}&\makecell{top-$K$ NDCG, top-$K$ MAP, pAUC, T-Push\\ P@$K$, R@$K$, pAp@$K$, P@R}&$\O\left( \max\left\{\frac{m}{B_1B_2\epsilon^4},\frac{1}{\min\{B_1,B_2\}\epsilon^4} \right\}\right)$\\ 
	     		\bottomrule
	\end{tabular}}
\end{table}

\subsection{Discussions and Open Problems}
We present a summary of different baseline algorithms and their properties in Table~\ref{tab:1}.   Next, we discuss the validity, significance and limitations of theoretical results of these algorithms. First, we can see that when $B_1=m=1$ (i.e., the number of blocks is one), the iteration complexity reduces to the standard one, i.e.,  $O(1/(B_2\epsilon^4))$, where $B_2$ is the mini-batch size. In addition, the scaling factor $m/B_1$ in the complexity term $O(\frac{m}{B_1B_2\epsilon^4})$ is reasonable since we only update variables for $B_1$ blocks out of $m$ blocks at each iteration. Second, the presented algorithms  are strong baseline algorithms. They are simple enough since they do not involve a nested loop for updating the the dual variable (in MBMMO) and the variable of the lower-level problem (in MBBO) with multiple steps at each iteration. They are practical because they do not require a large batch size as in many previous works yet still enjoy parallel speed-up using mini-batches. The update for the model parameter $\w$ is similar to the standard momentum and Adam-style update, which makes them easy to be implemented. In terms of limitations,  one (especially practitioners) might raise concerns/criticisms about the assumptions for the theoretical analysis, e.g., the smooth and Lipchitz continuity assumptions of $f_i, g_i$ for Algorithm~\ref{alg:mbbo}. While we understand such concerns, we would like to point out 
these conditions can hold in practice when the prediction score function $h_\w(\x)$ is Lipchitz continuous, smooth and bounded in terms of $\w$.

For future work,  we urge interested researchers in the community to further improve the performance of stochastic algorithms for solving EXM. There are several open questions to be considered for both theoretical and practical communities: (i) what is the optimal complexity for the considered problems for convex and non-convex objectives in both smooth and non-smooth settings?  (ii) how can we further accelerate the baseline algorithms by leveraging engineering insights and tricks?  (iii) how can we optimize these objectives in a distributed and federated setting? (iv) how can we analyze the  feature learning capability by optimizing X-risks for deep neural nets?

After the first version of this manuscript was finished, we have conducted studies to address some of the problems mentioned above. In~\cite{DBLP:conf/nips/JiangLW0Y22}, we have improved the complexities to $O(\max(\frac{1}{\min(B_1, B_2)\epsilon^4}, \frac{m}{B_1\sqrt{B_2}\epsilon^3}))$ for a non-convex objective by using advanced variance reduction techniques called MSVR. Its iteration complexity for a convex objective is $O(\max(\frac{1}{\min(B_1, B_2)\epsilon^2}, \frac{m}{B_1\sqrt{B_2}\epsilon^2}))$. However, both complexities do not exhibit the parallel speed-up in terms of the inner batch size $B_2$. Hu et al.~\cite{DBLP:journals/corr/abs-2310-03234} considered non-smooth FCCO problems where both $f$ and $g$ are weakly-convex and non-smooth. They utilized the MSVR estimator for estimating the inner function and obtained an iteration complexity of $O(\frac{m}{B_1\sqrt{B_2}\epsilon^6})$. He and Kasiviswanathan~\cite{DBLP:journals/corr/abs-2304-10613} have proposed stochastic extrapolation based estimator to reduce the bias of estimator $\nabla f(g)$ by using mini-batch samples of inner function $g$. Their algorithm improves the complexity of MSVR when $n\leq \Omega(\epsilon^{-2/3})$. However, their algorithm requires large inner and outer batch sizes and an extra condition that $f$ is third-order differntiable.   In a recent work~\cite{DBLP:conf/icml/Guo0LY23}, we developed a communication-efficient federated learning algorithm called FeDXL for solving the FCCO problem. If the data is evenly distributed over $K$ machines, its  iteration complexity is $O(\frac{m}{K\epsilon^4})$ and its communication complexity is $O(\frac{\sqrt{m}}{\sqrt{K}\epsilon^3})$. 

\section{Conclusions}
In this technical report, we have presented the algorithmic foundation of empirical X-risk minimization, including the formulations into different families of non-convex optimization problems, and their strong baseline stochastic optimization algorithms in our prior studies. We have also presented the convergence results of the baseline algorithms and discuss their applicability to different empirical X-risk minimization tasks. Finally, we present some discussions of the convergence results and future directions.

\bibliographystyle{plain}

\bibliography{AUC,All,Ref,Ref2,NDCG,All2,Draft,Ref3}

\end{document}